\documentclass[10pt,journal,compsoc]{IEEEtran}

\usepackage{array}
\usepackage[caption=false,font=normalsize,labelfont=sf,textfont=sf]{subfig}
\usepackage{textcomp}
\usepackage{stfloats}
\usepackage{url}
\usepackage{verbatim}
\usepackage{graphicx}
\usepackage{cite}

\usepackage{graphicx}
\usepackage{amsmath}
\usepackage{amssymb}
\usepackage{booktabs}
\usepackage[ruled,vlined,linesnumbered]{algorithm2e}
\usepackage{algpseudocode}
\usepackage{multirow}
\usepackage{ragged2e}
\renewcommand{\raggedright}{\leftskip=0pt \rightskip=0pt plus 0cm}
\DeclareMathOperator*{\argmax}{arg\,max}

\usepackage[pagebackref=true,breaklinks=true,colorlinks,bookmarks=false]{hyperref}
\usepackage[capitalize]{cleveref}
\crefname{section}{Sec.}{Secs.}
\Crefname{section}{Section}{Sections}
\Crefname{table}{Table}{Tables}
\crefname{table}{Tab.}{Tabs.}
\crefname{algocf}{alg.}{algs.}
\newcommand{\vect}[1]{{{\bf{#1}}}}

\usepackage{xspace}

\makeatletter
\DeclareRobustCommand\onedot{\futurelet\@let@token\@onedot}
\def\@onedot{\ifx\@let@token.\else.\null\fi\xspace}

\def\ie{\emph{i.e}\onedot}
\def\eg{\emph{e.g}\onedot}

\usepackage{color}

\begin{document}

\title{Post-train Black-box Defense via Bayesian Boundary Correction}

\author{He~Wang and Yunfeng~Diao
\IEEEcompsocitemizethanks{\IEEEcompsocthanksitem He Wang is with the University of College London, UK (e-mail: he\_wang@ucl.ac.uk, Corresponding Author)\\
\IEEEcompsocthanksitem Yunfeng Diao is with Hefei University of Technology, China (e-mail: diaoyunfeng@hfut.edu.cn) }\\
\thanks{He Wang
and Yunfeng Diao are co-first authors.}}

\markboth{Journal of \LaTeX\ Class Files,~Vol.~14, No.~8, August~2021}%
{Shell \MakeLowercase{\textit{et al.}}: A Sample Article Using IEEEtran.cls for IEEE Journals}


\IEEEtitleabstractindextext{
\raggedright{
\begin{abstract}
Classifiers based on deep neural networks are susceptible to adversarial attack, where the widely existing vulnerability has invoked the research in defending them from potential threats. Given a vulnerable classifier, existing defense methods are mostly white-box and often require re-training the victim under modified loss functions/training regimes. While the model/data/training specifics of the victim are usually unavailable to the user, re-training is unappealing, if not impossible for reasons such as limited computational resources. To this end, we propose a new post-train black-box defense framework. It can turn any pre-trained classifier into a resilient one with little knowledge of the model specifics. This is achieved by new joint Bayesian treatments on the clean data, the adversarial examples and the classifier, for maximizing their joint probability. It is further equipped with a new post-train strategy which keeps the victim intact, avoiding re-training. We name our framework Bayesian Boundary Correction (BBC). BBC is a general and flexible framework that can easily adapt to different data types. We instantiate BBC for image classification and skeleton-based human activity recognition, for both static and dynamic data. Exhaustive evaluation shows that BBC has superior robustness and can enhance robustness without severely hurting the clean accuracy, compared with existing defense methods. 
\end{abstract}}

\begin{IEEEkeywords}
Adversarial attack, Classification robustness, Human activity recognition
\end{IEEEkeywords}}
\maketitle
\section{Introduction}


\IEEEPARstart{D}{eep} neural network classifiers have been proven to be universally vulnerable to malicious perturbations on data and training, i.e. adversarial attack (AA)~\cite{GoodfellowFGSM, han2023interpreting}, causing alarming concerns because such perturbations are imperceptible to humans but destructive to machine intelligence. To this end, defense methods have emerged~\cite{han2023interpreting} where most research can be categorized into data enhancement and model enhancement. Data enhancement methods involve finding potential adversarial examples e.g. adversarial training (AT)~\cite{madry_towards_2018} and randomized smoothing (RS)~\cite{Cohen19_ICML}, or removing perturbation via denoising~\cite{nie2022diffusion}. The philosophy behind them is different. The former biases the classifier by exposing it to potential threats while the latter focuses on learning an accurate representation of the data distribution. Both can be seen as robustness enhancement via adding new training samples. However, the first kind diverts the classifier towards adversarial examples, and therefore often compromises the clean accuracy~\cite{TRADES}, and the second kind largely focuses on clean data while not explicitly considering the distribution of adversarial examples~\cite{nie2022diffusion}. In parallel, model enhancement methods explore particular architectures and training processes, e.g. adding regularization~\cite{fazlyab2024certified}, using robust loss functions~\cite{xu2023exploring}, employing tailored layers/activation~\cite{xulot}. Despite being effective in some settings, their overall performance is worse than data enhancement methods \cite{croce2020robustbench}. Recently, Bayesian treatment on model itself has shown great promise \cite{doan2022bayesian} but is still under-explored.

Given a pre-trained vulnerable classifier, most existing defenses can be regarded as \textit{white-box}, i.e. requiring the full knowledge of the model and robustly retraining the model, such as AT~\cite{madry_towards_2018}. However, \textit{white-box} defenses may not be applicable in some scenarios. From the perspective of developers, a model owner may refuse to share the model information for business or security considerations. From the perspective of end-users, large models are normally pre-trained on large datasets then shared. The end-users can directly use the pre-trained model or fine-tune it for downstream tasks. Retraining the model is computationally intensive and impractical. In addition, keeping the pre-trained model intact can also avoid undermining other tasks when the model is trained for multiple tasks. Last but not the least, robust retraining such as AT can severely compromise the benign accuracy~\cite{Yangcloserlook,TRADES}. Therefore, it is highly desirable to develop a \textit{post-train} \textit{black-box} defense, that does not require explicit knowledge of model architectures or parameters, nor necessitates re-training.



In this research, we aim for a general defense framework that is compatible with any pre-trained classifier, adaptive to incorporate domain-specific inductive biases, and lightweight incurring only a small additional cost, to achieve robustness. In other words, we do not aim for the most robust defense at any cost, but a cost-effective way for gaining robustness. Our key observation is that the different strengths in previous approaches, i.e. denoising methods emphasizing on the data manifold, AT and RS methods targeting at the adversarial distribution, and model enhancement methods aiming for the model robustness,  should be holistically considered. Therefore, we jointly model the data manifold, the adversarial distribution and the classifier, by learning the joint probability $p(\Tilde{\vect{x}}, \vect{x}, \vect{y}, \theta)$ where $\vect{x}$ and $\vect{y}$ are the data and labels, $\Tilde{\vect{x}}$ is the adversarial examples and $\theta$ is the classifier parameter. Through factorizing $p(\Tilde{\vect{x}}, \vect{x}, \vect{y}, \theta)$, we can explicitly represent all the key distributions including the data manifold $p(\vect{x}, \vect{y})$, the adversarial distribution $p(\Tilde{\vect{x}} | \vect{x}, \vect{y}, \theta)$ and the classifier distribution $p(\theta | \Tilde{\vect{x}}, \vect{x}, \vect{y})$. 


Specifically, we first parameterize the data distribution $p(\mathbf{x}, \mathbf{y})$ as an energy-based model~\cite{grathwohl_your_2020}. Observing that in AT different attackers can compute different adversarial examples for the same clean sample (i.e. one-to-many), we directly model $p(\Tilde{\mathbf{x}}, \mathbf{x}, \mathbf{y})$ where a flexible adversarial distribution $p(\Tilde{\mathbf{x}} | \mathbf{x}, \mathbf{y})$ can be obtained to capture the \textit{one-to-many} mapping between the clean and the adversarial samples. Finally, although there is an infinite number of classifiers that can perfectly classify data \cite{neal2012bayesian} (i.e. accuracy equivalency), we argue they have different robustness against different adversarial examples, i.e. a specific classifier can resist the attack from some adversarial examples but not others, or robustness inequivalency. This leads to a further generalization of $p(\Tilde{\mathbf{x}}, \mathbf{x}, \mathbf{y})$ into $p(\Tilde{\vect{x}}, \vect{x}, \vect{y}, \theta)$, where the classifier posterior $p(\theta | \Tilde{\vect{x}}, \vect{x}, \vect{y})$ considers all robust classifiers. We name our method Bayesian Boundary Correction (BBC).

BBC is straightforward but effective, with key novelties shown in \cref{tab:comparison_methods}. The first novelty is the simultaneous Bayesian treatments on the data manifold, the adversarial distribution and the classifier, inspired by the clean-adversarial one-to-many mapping and the robustness inequivalency of classifiers. The second novelty is our new post-train defense strategy to simultaneously tackle the sampling difficulties in inference on Bayesian Neural Networks and keep the victim intact. BBC appends a small Bayesian component to the victim, which makes BBC applicable to pre-trained classifiers and requires minimal prior knowledge of them, achieving post-train black-box defense.

As a result, BBC can turn pre-trained classifiers into resilient ones. It also circumvents model re-training, avoids heavy memory footprint and speeds up adversarial training. BBC leads to a general defense mechanism against a variety of attackers which are not known \textit{a priori} during training. We evaluate BBC on a large number of classifiers on image and skeletal motion data, and compare it with existing methods. Empirically, BBC can effectively boost the robustness of classifiers against attack, and does not severely sacrifice accuracy for robustness, as opposed to the common observation of such trade-off in other methods~\cite{Yangcloserlook}.

Formally, we propose: 1) a new general and adaptive post-train black-box defense framework. 2) a new Bayesian perspective on a joint distribution of clean data, adversarial examples and classifiers. 3) a new flexible way of incorporating domain-specific inductive biases for robust learning. 4) a new post-train Bayesian strategy to keep the blackboxness of classifiers and avoid heavy memory footprint.

Our paper is an extension of \cite{BEAT}. In this journal extension, we have 1) extended the previous work to a more general defense framework compatible with any pre-trained classifier and adaptive to different tasks, with new experiments and comparisons in image classification; 2) proposed two new variants of BBC against AutoAttack, added new evaluation for gradient obfuscation and more comprehensive ablation studies; 3) added more detailed derivation of the model and the inference method, included a broader and deeper literature review and conducted more extensive analysis and discussion.


\begin{table}[!htb]
\vspace{-0.3cm}
\caption{A high-level comparison between our method and existing methods. DM: data manifold. AD: adversarial distribution. CD: classifier distribution. PT: post-train.}\label{tab:comparison_methods}%
\centering
\setlength{\tabcolsep}{1 mm}{
\begin{tabular}{ccccc}
\toprule
Method & DM & AD & CD & PT\\
\midrule
AT~\cite{madry_towards_2018} & n/a & point estimation & n/a & no\\
RS~\cite{Cohen19_ICML} & n/a & simplified & n/a & no\\
Denoising~\cite{nie2022diffusion} & Yes & n/a & n/a & no\\
Model enhancement~\cite{xulot}& simplified & simplified & n/a & some\\
Bayesian~\cite{doan2022bayesian} & n/a & simplified & Yes & no\\
\midrule
Ours & Yes & Yes & Yes & Yes\\
\bottomrule
\end{tabular}}
\vspace{-0.5cm}
\end{table}

\section{Related Work}
\noindent{\textbf{Adversarial Attack.}} Since identifying the vulnerability in deep learning models for the first time~\cite{GoodfellowFGSM}, the community has developed numerous adversarial attacks~\cite{han2023interpreting}. Many of these attacks involve computing or estimating the gradient of the model to generate adversarial examples under both the white-box and black-box settings~\cite{GoodfellowFGSM,madry_towards_2018,Bandits,ma2023transferable}. Later, the discovery of the gradient obfuscation phenomenon prompts the development of adaptive attacks~\cite{athalye2018obfuscated,tramer2020adaptive}. While static data has attracted most of the attention, the attack on time-series data has recently emerged \eg in general time-series analysis~\cite{karim_adversarial_2020} and video understanding~\cite{wei2022cross,zhang2024vulnerabilities}. Unlike static data, time-series data contains rich dynamics, which makes it difficult to adapt generic methods~\cite{madry_towards_2018,BA}. Therefore, time-series attacks need to be carefully designed for specific data types~\cite{wang_understanding_2021}. Very recently, skeleton-based Human Activity Recognition (S-HAR) classifiers, one active sub-field in dynamic data, has been shown to be extremely vulnerable~\cite{liu_adversarial_2020,zheng_towards_2020, tanaka2021adversarial,diao_basarblack-box_2021, BASAR_journal}, calling for mitigation.

\noindent{\textbf{Adversarial Training (AT).}} AT methods~\cite{GoodfellowFGSM, madry_towards_2018} are among the most effective defense techniques to date. Recently, different training strategies have been designed to significantly improve the vanilla AT~\cite{AWP,LAS-AT,rat}. AWP~\cite{AWP} enhances robustness by regularizing the flatness of the weight loss landscape. LAS-AT~\cite{LAS-AT} explores a dynamic adversary sampling in AT. RAT~\cite{rat} adds random noise into robust deterministic weights. While effective in improving robustness, AT forces the classifier to correctly classify the most aggressive adversarial example during training at the cost of clean accuracy~\cite{madry_towards_2018,MART}, which is regarded as an inherent trade-off between clean accuracy and adversarial robustness~\cite{Yangcloserlook,TRADES}. More recent attempts to alleviate the accuracy-robustness trade-off include early-stopping version of AT~\cite{zhang2020attacks}, studying the manifold of adversaries and clean data~\cite{stutz2019disentangling,BASAR_journal}, using natural classifier boundary guidance~\cite{LBGAT}. A series of work~\cite{gowal2020uncovering,WangPDL0Y23} suggest that the robust generalization can be remarkably improved by largely using extra data in AT. Despite these advancements, this problem remains far from being solved, as there still exists an obvious gap between accuracy and robustness. We revisit this issue in our paper and narrow the gap even further, without requiring extra data and retraining the model. 

So far, most AT studies have focused on static data, e.g. image data, leaving defense for time-series data less explored. In this paper, we further propose a general defense framework that can adapt to different data types, including static image data and time-series motion data.

\noindent{\textbf{Bayesian Defense.}} Bayesian Neural Networks (BNNs) with the capabilities of modeling uncertainty have shown great promise in defending against adversarial examples~\cite{AdvBNN}. Carbone et al.~\cite{carbone2020robustness} theoretically demonstrate that BNNs are robust to gradient-based adversarial attacks in the over-parameterization and large data limit. Ye et al.~\cite{ye2018bayesian} propose Bayesian adversarial learning to incorporate the uncertainty of data and model parameters. Adv-BNN~\cite{AdvBNN} scales Bayesian adversarial training to more complex data and adds randomness to all weights in the network. Further, IG-BNN~\cite{doan2022bayesian} employs Stein Variational Gradient Descent~\cite{liu2016stein} for better sampling the posterior distribution, and to maintain the same measure of information content learned from the given input and its adversarial counterpart. We further extend the Bayesian 
defense family by introducing a new post-train Bayesian strategy that enables fast Bayesian training under the black-box defense setting.


\noindent{\textbf{Black-box Defense.}} Black-box or post-processing defense has been largely unexplored, despite~\cite{zhang2022robustify,chenadversarial,BEAT}. Zhang et al.~\cite{zhang2022robustify} study the certified black-box defense from a zeroth-order optimization perspective. Chen et al.~\cite{chenadversarial} propose to perturb the DNN's output scores to fool the attackers, but such a defense strategy is only effective for preventing score-based query attacks and is not applicable to other types of attacks such as gradient-based attacks. Wang et al.~\cite{BEAT} propose the first black-box defense for S-HAR. In this paper, we propose a new general black-box defense framework, adaptive to different data types. We empirically demonstrate that BBC can defend both white-box and black-box attacks without sacrificing accuracy.

\section{Methodology}
\label{sec:ebm}
Given data $\mathbf{x} \in \mathbf{X}$ and label $y\in\mathbf{y}$, a classifier can be seen as as an energy-based mode:  $p_{\theta}(\mathbf{x}, y) = exp(g_{\theta}(\mathbf{x})[y])/ Z(\theta)$ parameterized by $\theta$~\cite{grathwohl_your_2020}. Here $p_{\theta}(\mathbf{x}, y) = p_{\theta}(y\vert\mathbf{x})p_{\theta}(\mathbf{x})$. $p_{\theta}(y\vert\mathbf{x})$ is what classifiers maximize, and $p_{\theta}(\mathbf{x})$ can be parameterized by an energy function:
\begin{equation}
\label{eq:dataDis}
    p_{\theta}(\mathbf{x}) = \frac{exp(-E_{\theta}(\mathbf{x}))}{Z(\theta)} = \frac{\sum_{y\in\mathbf{y}} exp(g_{\theta}(\mathbf{x})[y])}{Z(\theta)}
\end{equation}
where $E_{\theta}$ is an energy function parameterized by $\theta$, $Z(\theta) = \int_{\mathbf x} exp(-E_\theta({\mathbf x}))d{\mathbf x}$ is a normalizing constant. This energy-based interpretation allows an arbitrary $E_{\theta}$ to describe a continuous density function, as long as it assigns low energy values to observations and high energy everywhere else. A general choice for $E_{\theta}$ is an exponential function in \cref{eq:dataDis}, where $g_{\theta}$ is a classifier and $g_{\theta}(\mathbf{x})[y]$ gives the $y$th logit for class $y$. $\theta$ can be learned via maximizing $p_{\theta}(\mathbf{x}, y)$:
\begin{align}
\label{eq:jointEnergy}
    &log\ p_{\theta}(\mathbf{x}, y) = log\ p_{\theta}(y\vert \mathbf{x}) + log\ p_{\theta}(\mathbf{x}) \nonumber \text{  where} \\
    &p_{\theta}(y\vert\mathbf{x}) = \frac{p_{\theta}(\mathbf{x},y)}{p_{\theta}(\mathbf{x})} = \frac{exp(g_{\theta}(\mathbf{x})[y])}{\sum_{y'\in\mathbf{y}} exp(g_{\theta}(\mathbf{x})[y'])}
\end{align}
Compared with only maximizing $log\ p(y\vert \mathbf{x})$ as discriminative classifiers do, maximizing $log\  p(\mathbf{x}, y)$ can provide many benefits such as good accuracy, robustness and out-of-distribution detection~\cite{grathwohl_your_2020}.

\subsection{Joint Distribution of Data and Adversaries}\label{sec:jointModel}
A robust classifier is defined as $g_\theta(\mathbf{x}) = g_\theta(\Tilde{\mathbf{x}}) \text{ where}\ \Tilde{\mathbf{x}} = \mathbf{x} + \mathbf{\delta}, \mathbf{\delta}\in\pi$, where $\mathbf{\delta}$ is drawn from some perturbation set $\pi$, computed by an attacker. Here $g_\theta$ needs to capture the whole adversarial distribution to be able to resist potential attacks post-train, which is non-trivial since the attacker is not known \textit{a priori}. This has led to two strategies: defending against the most adversarial sample from an attacker (a.k.a Adversarial Training or AT~\cite{madry_towards_2018}) or training on data with noises (a.k.a Randomized Smoothing or RS~\cite{Cohen19_ICML}). However, in addition to requiring re-training, both approaches lead to a trade-off between accuracy and robustness~\cite{Yangcloserlook}. Also neither can fully capture the adversarial distribution.

We start from a straightforward yet key conceptual deviation from literature~\cite{han2023interpreting}: assuming there is an adversarial distribution over all adversarial examples which could be computed by all possible attackers. This assumption is driven by the observation that different attackers in AT might compute different adversarial examples even for the same clean example, indicating that there is a distribution of adversarial examples given one clean example. Further, all adversarial examples are close to the clean data~\cite{stutz2019disentangling, diao_basarblack-box_2021}. So for a vulnerable classifier, they also have relatively low energy. Therefore, we add the adversarial samples $\Tilde{\mathbf{x}}$ to the joint distribution $p(\mathbf{x}, \Tilde{\mathbf{x}}, y)$, and then further extend it into a new clean-adversarial energy-based model:
\begin{equation}
\label{eq:fullJointEnergy}
    p_{\theta}(\mathbf{x}, \Tilde{\mathbf{x}}, y) = \frac{exp\{g_{\theta}(\mathbf{x})[y] + g_{\theta}(\mathbf{\Tilde{\mathbf{x}}})[y] - \lambda d(\mathbf{x}, \Tilde{\mathbf{x}})\}}{\Tilde{Z}(\theta)}
\end{equation}
where $\Tilde{Z}(\theta) = \int_{\hat{\mathbf x}} exp(-E_\theta({\hat{\mathbf x}}))d{\hat{\mathbf x}}$, $\hat{\mathbf x}\in \mathbf{X} \cup \mathbf{X}'$ is either a clean or an adversarial sample. $\mathbf{x}$ and $\Tilde{\mathbf{x}}$ are the clean examples and their adversaries under class $y$. $\lambda$ is a weight and $d(\mathbf{x}, \Tilde{\mathbf{x}})$ measures the distance between the clean samples and their adversaries. \cref{eq:fullJointEnergy} bears two assumptions. First, adversaries are in the low-energy (high-density) area as they are very similar to data. Also, their energy should increase (or density should decrease) when they deviate away from the clean samples, governed by $d(\mathbf{x}, \Tilde{\mathbf{x}})$.

In $p_{\theta}(\mathbf{x}, \Tilde{\mathbf{x}}, y) = p_{\theta}(\Tilde{\mathbf{x}} \vert \mathbf{x}, y)p_{\theta}(\mathbf{x}, y)$, 
$p_{\theta}(\Tilde{\mathbf{x}} \vert \mathbf{x}, y)$ is a new term. To further understand this term, for each $\mathbf{x}$, we take a Bayesian perspective and assume there is a distribution of adversarial samples $\Tilde{\mathbf{x}}$ around $\mathbf{x}$. This is reasonable as every adversarial sample can be traced back to a clean sample (\ie every adversarial example should be visually similar to a clean example), and there is a \textit{one-to-many} mapping from the clean samples to the adversarial samples. Then $p_{\theta}(\Tilde{\mathbf{x}} \vert \mathbf{x}, y)$ is a \textit{full Bayesian treatment} of all adversarial samples:
\begin{align}
\label{eq:adDist}
    p&_{\theta}(\Tilde{\mathbf{x}} \vert \mathbf{x}, y) = \frac{p_{\theta}(\mathbf{x}, \Tilde{\mathbf{x}}, y)}{p_{\theta}(\mathbf{x}, y)}\nonumber \\
    &=\frac{exp\{g_{\theta}(\mathbf{x})[y] + g_{\theta}(\mathbf{\Tilde{\mathbf{x}}})[y] - \lambda d(\mathbf{x}, \Tilde{\mathbf{x}})\}}{exp(g_{\theta}(\mathbf{x})[y])} \frac{Z(\theta)}{\Tilde{Z}(\theta)} \nonumber \\
    &\approx exp\{g_{\theta}(\mathbf{\Tilde{\mathbf{x}}})[y] - \lambda d(\mathbf{x}, \Tilde{\mathbf{x}})\}     
\end{align}
where we use approximation $Z(\theta)/\Tilde{Z}(\theta) \approx 1$ when $\delta$ is small, as $\displaystyle{\lim_{\delta\rightarrow 0}}  [Z(\theta)/\Tilde{Z}(\theta)]=1$. \cref{eq:adDist} ensures that adversarial samples will be given low energy and thus high density, so that $g_{\theta}$ is now capable of taking the adversarial distribution into consideration during training.

\subsubsection{Connections to Existing Defense Methods}
\label{sec:connection}
BBC has intrinsic connections with AT and RS. Since the potential attacker is unknown \textit{a priori}, $d(\mathbf{x}, \Tilde{\mathbf{x}})$ in Eq.~\ref{eq:fullJointEnergy} needs to capture the full adversarial distribution. In this sense, properly instantiating $d(\mathbf{x}, \Tilde{\mathbf{x}})$ can recover both AT and RS. To see this, AT solves~\cite{madry_towards_2018}:
\begin{equation}
\min_{\theta} E_{\mathbf x}[\max_{\delta\in \pi} L(\theta, \mathbf{x} + \delta, y)],     
\end{equation}
where $L$ is the classification loss, the perturbation set $\pi$ is constrained within a ball, and $\delta$ needs to be computed by a pre-defined attacker. Here, a basic AT is recovered when $d(\mathbf{x}, \Tilde{\mathbf{x}})$ is the Euclidean distance within the ball $\pi$. In RS, the robust classifier is obtained through~\cite{Cohen19_ICML}: 
\begin{equation}
    \argmax_{y \in \mathbf{y}} p(g_\theta(\mathbf{x} + \delta)=y) \text{ where } \delta \sim \mathbf{N}
\end{equation}
where $\pi$ is an isotropic Gaussian. $d(\mathbf{x}, \Tilde{\mathbf{x}})$ essentially plays the role of the Gaussian to describe the perturbation set. 

However, neither AT nor RS capture the fine-grained structure of the adversarial distribution, because AT merely uses the most aggressive adversarial example, and RS often employs simple \textit{isotropic} distributions (\eg Gaussian/Laplacian)~\cite{Cohen19_ICML}. Therefore, we argue $d(\mathbf{x}, \Tilde{\mathbf{x}})$ should be data/task specific and not be restricted to isotropic forms. This is because adversarial samples are near the data manifold, both on-manifold and off-manifold~\cite{stutz2019disentangling, diao_basarblack-box_2021}, so the data manifold geometry should dictate the parameterization of $d(\mathbf{x}, \Tilde{\mathbf{x}})$. A proper $d(\mathbf{x}, \Tilde{\mathbf{x}})$ allows us to expand the space around the manifold in a potentially non-isotropic way, like adding a `thickness' to the data manifold, which can be achieved implicitly or explicitly. Explicit formulations can be used if it is straightforward to parameterize the manifold geometry; or a data-driven model can be used to implicitly learn the manifold. Either way, the manifold can then be devised with a distance function to instantiate $d(\mathbf{x}, \Tilde{\mathbf{x}})$. 

\subsection{Bayesian Classifier for Further Robustness}
\label{sec:bayesianClassifier}
Although \cref{eq:fullJointEnergy} considers the full distribution of the data and the adversarial examples, it is still a \textit{point estimation} with respect to the model $\theta$. From a Bayesian perspective, there is a distribution of models which can correctly classify $\mathbf{x}$, \ie there is an infinite number of ways to draw the classification boundaries (accuracy equivalency). Our insight is these models can vary in terms of their robustness (robustness in-equivalency). Intuitively, a single boundary can be robust against certain adversaries, \eg the distance between the boundary and some clean data examples are large hence requiring larger perturbations for attack. However, they might not be similarly robust against other adversaries. A collection of all boundaries can be more robust because they provide different between-class distances~\cite{Yangcloserlook} and local boundary continuity~\cite{xu2023exploring}, which are all good sources for robustness. Therefore, we augment~\cref{eq:fullJointEnergy} to incorporate the network weights $\theta$: $p(\theta, \mathbf{x}, \Tilde{\mathbf{x}}, y) = p(\mathbf{x}, \Tilde{\mathbf{x}}, y \vert \theta) p(\theta)$, 
where $p(\mathbf{x}, \Tilde{\mathbf{x}}, y \vert \theta)$ is essentially~\cref{eq:fullJointEnergy} and $p(\theta)$ is the prior of network weights. 

Now we have a new Bayesian joint model of clean data, adversarial examples and the classifier. From the Bayesian perspective, maximizing \cref{eq:fullJointEnergy} is equivalent to using a flat $p(\theta)$ and applying iterative \textit{Maximum a posteriori} (MAP) optimization. However, even with a flat prior, a MAP optimization is still a point estimation on the model, and cannot fully utilize the full posterior distribution~\cite{saatci_bayesian_2017}. In contrast, we propose to use \textit{Bayesian Model Averaging}:
\begin{align}
\label{eq:bayesianClassifier}
    p(y' \vert \mathbf{x}', \mathbf{x}, \Tilde{\mathbf{x}}, y) = E_{\theta\sim p(\theta)}[p(y' \vert \mathbf{x}', \mathbf{x}, \Tilde{\mathbf{x}}, y, \theta)] \nonumber \\
    \approx \frac{1}{N}\sum_{i = 1}^Np(y'\vert\mathbf{x}',\theta_i), \theta \sim p(\theta \vert \mathbf{x}, \Tilde{\mathbf{x}}, y)
\end{align}
where $\mathbf{x}'$ and $y'$ are a new sample and its predicted label, $p(\theta)$ is a flat prior, $N$ is the number of models. We expect such a Bayesian classifier to be more robust against attack while achieving good accuracy, because models from the high probability regions of $p( \theta \vert \mathbf{x}, \Tilde{\mathbf{x}}, y)$ provide both. This is vital as we do not know the attacker in advance. To train such a classifier, the posterior distribution $p( \theta \vert \mathbf{x}, \Tilde{\mathbf{x}}, y)$ needs to be sampled as it is intractable.

\subsubsection{Necessity of a Post-train Bayesian Strategy}
Unfortunately, it is not straightforward to design such a Bayesian treatment (\cref{eq:bayesianClassifier}) on pre-trained classifiers due to several factors. First, sampling the posterior distribution $p( \theta \vert \mathbf{x}, \Tilde{\mathbf{x}}, y)$ is prohibitively slow. Considering the large number of parameters in classifiers (possibly over several million), sampling would mix extremely slowly in such a high dimensional space (if at all). 

In addition, to end-users, large models are normally pre-trained on large datasets then shared. The end-users can fine-tune or directly use the pre-trained model. It is not desirable/possible to re-train the models. This can be because the owner of the model refuses to share the model details, or they cannot share the training data for security/ethical reasons, or simply the end-users do not have necessary computing capacity to retrain the model. 

Finally, most classifiers consist of two parts: feature extraction and boundary computation. The data is pulled through the first part to be mapped into a latent feature space, then the boundary is computed, \eg through fully-connected layers. The feature extraction component is well learned in the pre-trained model. Keeping the features intact can avoid re-training the model, and avoid undermining other tasks when the features are learned for multiple tasks, \eg under representation/self-supervised learning.

Therefore, we propose a \textit{post-train} Bayesian strategy for \textit{black-box} defense. We keep the pre-trained classifier intact and append a tiny model with parameters $\theta'$ behind the classifier using a skip connection: $g_{\theta'}(\mathbf{x}) = f_{\theta'}(\phi(\mathbf{x})) + g_\theta(\mathbf{x})$, in contrast to the original logits = $g_\theta(\mathbf{x})$. $\phi(\mathbf{x})$ can be the latent features of $\mathbf{x}$ or the original logits $\phi(\mathbf{x}) = g_\theta(\mathbf{x})$. We employ the latter setting based on the ablation studies in \cref{sec:ablation} and to keep the \textit{blackboxness} of BBC. We can replace all the $g_\theta(\mathbf{x})$ above with $g_{\theta'}(\mathbf{x})$ now. \cref{eq:bayesianClassifier} then becomes:
\begin{align}
\label{eq:ptBayesianClassifier}
    p(y' \vert \mathbf{x}', \mathbf{x}, \Tilde{\mathbf{x}}, y) = E_{\theta'\sim p(\theta')}[p(y' \vert \mathbf{x}', \mathbf{x}, \Tilde{\mathbf{x}}, y, \theta, \theta')] \nonumber \\
    \approx \frac{1}{N}\sum_{i = 1}^Np(y'\vert \mathbf{x}',\theta'_i,\theta), \theta' \sim p(\theta' \vert \mathbf{x}, \Tilde{\mathbf{x}}, y, \theta)
\end{align}
where $\theta$ is fixed after pre-training. Then BBC training can be conducted through alternative sampling:
\begin{align}
\label{eq:ptModelSampling}
    &\{\mathbf{x}, \Tilde{\mathbf{x}}, y\}_t \vert \theta, \theta'_{t-1} \sim p(\mathbf{x}, \Tilde{\mathbf{x}}, y \vert \theta, \theta'_{t-1}) \nonumber\\
    &\theta'_t \vert \{\mathbf{x}, \Tilde{\mathbf{x}}, y\}_t, \theta \sim p(\theta' \vert \{\mathbf{x}, \Tilde{\mathbf{x}}, y\}_t, \theta)
\end{align}
Although $f_{\theta'}$ can be any model, surprisingly a simple two-layer fully-connected layer network (with the same dimension as the original output) proves to work well in all cases. During attack, we attack the full model $g_{\theta'}$. We use $N = 5$ models in all experiments and explain the reason in the ablation study later. Overall, since $f_{\theta'}$ is much smaller than $g_{\theta}$, training $g_{\theta'}$ is faster than re-training $g_{\theta}$.  


\subsection{Inference on BBC}
\label{sec:inference}
Following \cref{eq:ptModelSampling}, we sample $\theta'$ by Stochastic Gradient Hamiltonian Monte Carlo~\cite{chen_stochastic_2014}. However, it cannot efficiently explore the target density due to the high correlations between parameters in $\theta'$. Therefore, we use Stochastic Gradient Adaptive Hamiltonian Monte Carlo~\cite{springenberg_bayesian_2016}:
\begin{align}
\label{eq:SGAHMC}
&\theta'_{t+1} = \theta'_t - \sigma^2\mathbf{C}^{-1/2}_{\theta'_t}\mathbf{h}_{\theta'_t} + \mathbf{N}(0, 2F\sigma^3\mathbf{C}^{-1}_{\theta'_t} - \sigma^4\mathbf{I}) \nonumber\\
&\mathbf{C}_{\theta'_t} \leftarrow (1 - \tau^{-1})\mathbf{C}_{\theta'_t} + \tau^{-1}\mathbf{h}_{\theta'_t}^2
\end{align}
where $\sigma$ is the step size, $F$ is called friction coefficient, $\mathbf{h}$ is the stochastic gradient of the system, $\mathbf{N}$ is a Normal distribution and $\mathbf{I}$ is an identity matrix, $\mathbf{C}$ is a pre-conditioner and updated by an exponential moving average and $\tau$ is chosen automatically~\cite{springenberg_bayesian_2016}.

Starting from some initial $\theta'_0$, iteratively using \cref{eq:SGAHMC} will lead to different $\theta'_{t+1}$s that are all samples from the posterior $p(\theta' \vert \{\mathbf{x}, \Tilde{\mathbf{x}}, y\}_t, \theta)$. The key component in \cref{eq:SGAHMC} is the gradient of the system $\mathbf{h}$. For BBC, the natural option is to follow the gradient that maximizes the log-likelihood of the joint probability (\cref{eq:fullJointEnergy}):
\begin{align}
\label{eq:logLikelihood}
    &log\ p_{\theta'}(\mathbf{x}, \Tilde{\mathbf{x}}, y) = log\  p_{\theta'}(\Tilde{\mathbf{x}} \vert \mathbf{x}, y) + log\ p_{\theta'}(\mathbf{x}, y) \nonumber \\
    &= log\  p_{\theta'}(\Tilde{\mathbf{x}} \vert \mathbf{x}, y) + log\ p_{\theta'}(y\vert\mathbf{x}) + log\ p_{\theta'}(\mathbf{x}) 
\end{align}
where $y$ is randomly sampled and $log\ p_{\theta'}(y\vert\mathbf{x})$ is simply a classification likelihood and can be estimated via \eg cross-entropy. Both $p_{\theta'}(\mathbf{x})$ and $p_{\theta'}(\Tilde{\mathbf{x}} \vert \mathbf{x}, y)$ are intractable, so sampling is needed. 

To computer $\mathbf{h}$ from \Cref{eq:logLikelihood}, we need to compute three gradients $\frac{\partial log p_{\theta'}(\Tilde{\mathbf{x}} | \mathbf{x}, y)}{\partial\theta'}$, $\frac{\partial log p_{\theta'}(y|\mathbf{x})}{\partial\theta'}$ and $\frac{\partial log p_{\theta'}(\mathbf{x})}{\partial\theta'}$. Instead of directly maximizing $log p_{\theta'}(y|\mathbf{x})$, we minimize the \textit{cross-entropy} on the logits, so $\frac{\partial log p_{\theta'}(y|\mathbf{x})}{\partial\theta'}$ is straightforward. Next, $\frac{\partial log p_{\theta'}(\mathbf{x})}{\partial\theta'}$ can be approximated by~\cite{nijkamp_anatomy_2019}:
\begin{equation}
\label{eq:pxgradient}
    \frac{\partial log p_{\theta'}(\mathbf{x})}{\partial\theta'} \approx \frac{\partial}{\partial\theta'}[\frac{1}{L_1}\sum_{i=1}^{L_1}U(g_{\theta'}(\mathbf{x}^+_i)) - \frac{1}{L_2}\sum_{i=1}^{L_2}U(g_{\theta'}(\mathbf{x}^-_i))]
\end{equation}
where $U$ gives the mean over the logits, \{$\mathbf{x}^+_i$\}$_{i=1}^{L_1}$ are a batch of training samples and \{$\mathbf{x}^-_i$\}$_{i=1}^{L_2}$ are i.i.d. samples from $p_{\theta'}(\mathbf{x})$ via Stochastic Gradient Langevin Dynamics (SGLD)~\cite{welling_bayesian_2011}:
\begin{align}
\label{eq:SGLD_x}
    \mathbf{x}^-_{t+1} = \mathbf{x}^-_{t} + \frac{\epsilon^2}{2} \frac{\partial log p_{\theta'}(\mathbf{x}^-_t)}{\partial \mathbf{x}^-_t} + \epsilon E_t, \epsilon > 0, E_t\in \mathbf{N}(0, \mathbf{I}) 
\end{align}
where $\epsilon$ is a step size, $\mathbf{N}$ is a Normal distribution and $\mathbf{I}$ is an identity matrix. Similarly for $\frac{\partial log p_{\theta'}(\Tilde{\mathbf{x}} | \mathbf{x}, y)}{\partial\theta'}$:
\begin{align}
\label{eq:pxtildegradient}
    \frac{\partial log p_{\theta'}(\Tilde{\mathbf{x}}|\mathbf{x}, y)}{\partial\theta'} = \frac{\partial}{\partial\theta'}\{g_{\theta'}(\mathbf{\Tilde{\mathbf{x}}})[y] - \lambda d(\mathbf{x}, \Tilde{\mathbf{x}})\} 
\end{align}
where $\Tilde{\mathbf{x}}$ can be sampled via:
\begin{equation}
\label{eq:SGLD_xtilde}
    \Tilde{\mathbf{x}}_{t+1} = \Tilde{\mathbf{x}}_{t} + \frac{\epsilon^2}{2} \frac{\partial log p_{\theta'}(\Tilde{\mathbf{x}}_t| \mathbf{x}, y)}{\partial\Tilde{\mathbf{x}}_t} + \epsilon E_t, \epsilon > 0, E_t\in \mathbf{N}(0, \mathbf{I})
\end{equation}
Further, instead of naive SGLD, we use Persistent Contrastive Divergence~\cite{tieleman_training_2008} with a random start. The BBC inference is detailed in \cref{alg:BBC}.

\begin{algorithm}[tb]
\SetAlgoLined
\textbf{Input}: $\mathbf{x}$: training data; $N_{tra}$: the number of training iterations; $M_1$ and $M_2$: sampling iterations; $M_{\theta'}$: sampling iterations for $\theta'$; $f_{\theta'}$: appended models with parameter $\{\theta'_1, \dots, \theta'_N\}$; $N$: the number of appended models\;
\textbf{Output}: $\{\theta'_1, \dots, \theta'_N\}$: appended network weights\;
\textbf{Init}: randomly initialize $\{\theta'_1, \dots, \theta'_N\}$\;

\For{i = 1 to $N_{tra}$}{
    \For{n = 1 to $N$}{
        Randomly sample a mini-batch data $\{\mathbf{x}, y\}_i$\;
        Compute $h_1 = \frac{\partial log p_{\theta'}(y|\mathbf{x})}{\partial\theta'}$\;
        Obtain $\mathbf{x}_0$ via random noise~\cite{du_implicit_2020}\;
        for t = 1 to $M_1$, sample $\mathbf{x}_t$ via \cref{eq:SGLD_x}\;
        Compute $h_2 = \frac{\partial log p_{\theta'}(\mathbf{x})}{\partial\theta'}$ via \cref{eq:pxgradient}\;
        Obtain $\Tilde{\mathbf{x}}_0$ from $\mathbf{x}_i$ with a perturbation\;
        for t = 1 to $M_2$, sample $\Tilde{\mathbf{x}}_t$ via \cref{eq:SGLD_xtilde}\;
        Compute $h_3 = \frac{\partial log p_{\theta'}(\Tilde{\mathbf{x}} | \mathbf{x}, y)}{\partial\theta'}$ via \cref{eq:pxtildegradient}\;
        $\mathbf{h}_{\theta'} = h_1 + h_2 + h_3$\;
        for t = 1 to $M_{\theta'}$, update $\theta'_n$ via \cref{eq:SGAHMC}\;
    }
}
\Return $\{\theta'_1, \dots, \theta'_N\}$\;
\caption{Inference on BBC}
\label{alg:BBC}
\end{algorithm}

\subsection{Instantiating BBC for Different Data and Tasks}
\label{sec:instantiate}
We instantiate BBC for two types of widely investigated data: image data in image classification and skeleton motion data in S-HAR. The former is a general classification task which has been heavily studied, while the latter is time-series data which contains rich dynamics. Given the flexibility of BBC, the instantiation can be achieved by only specifying $d(\mathbf{x}, \Tilde{\mathbf{x}})$ in \cref{eq:adDist}, where we introduce domain-specific inductive biases. Our general idea is that $d(\mathbf{x}, \Tilde{\mathbf{x}})$ should faithfully reflect the distance of any sample from the underlying data manifold, as discussed in \cref{sec:connection}.

\subsubsection{Perceptual Distance for BBC in Images Classification}
The adversarial samples are imperceptible to humans and hence are distributed closely to the data manifold. This suggests that the data manifold being highly tuned with human visual perception, such that the data manifold can be accurately described by the true perceptual distance~\cite{zhang2018unreasonable}. However, the true perceptual distance cannot be directly computed for image data. Considering that the perceptual similarity can be intuitively linked to the deep visual representation~\cite{zhang2018unreasonable}, we propose to use neural perceptual distance, i.e. a neural network-based approximation of the true perceptual distance. Specifically, we define $d(\mathbf{x}, \Tilde{\mathbf{x}})$ via LPIPS~\cite{zhang2018unreasonable} which trains a network $h(\cdot)$ to compute the feature distances. We first extract a feature stack from $L$ layers in $h(\cdot)$ and normalize them across the channel dimension. With $\hat{h}_l(\mathbf{x})$ denoting the internal channel-normalized activation at the $l$-th layer, $\hat{h}_l(\mathbf{x})$ is normalized again by the layer size, so $d(\mathbf{x}, \Tilde{\mathbf{x}})$ is defined as:
\begin{align}
\label{eq:perception}
    d(\mathbf{x}, \Tilde{\mathbf{x}}) = \sum_l^L w_l ||\frac{\hat{h}_l(\mathbf{x})}{\sqrt{W_lH_l}} - \frac{\hat{h}_l(\Tilde{\mathbf{x}})}{\sqrt{W_lH_l}}||^2_2 
\end{align}
where $W_l$ and the $H_l$ are the width and height of the activations in layers $l$, $w_l$ is the weights. This distance function helps to quantify the perceptual similarity between a clean sample and its adversarial counterpart $\mathbf{x}$, so \cref{eq:perception} describes the adversarial distribution near the image manifold. In theory, any good representation learner can be used as $h(\cdot)$. In practice, we find applying the AlexNet pre-trained with ImageNet as $h(\cdot)$ can achieve a good robust performance.

\subsubsection{Natural Motion Manifold for BBC in S-HAR}
The motion manifold is well described by the motion dynamics and bone lengths~\cite{diao_basarblack-box_2021}. Therefore, we design $d(\mathbf{x}, \Tilde{\mathbf{x}})$ so that the energy function in \cref{eq:adDist} also assigns low energy values to the adversarial samples bearing similar motion dynamics and bone lengths:
\begin{align}
\label{eq:simMeasure}
    d(\mathbf{x}, \Tilde{\mathbf{x}}) = \frac{1}{MB}\sum||BL(\mathbf{x}) - BL(\Tilde{\mathbf{x}})||^2_p \nonumber \\
    + \frac{1}{MJ}\sum||q_{m,j}^k(\mathbf{x}) - \Tilde{q}^k_{m,j}(\Tilde{\mathbf{x}})||^2_p
\end{align}
where $\mathbf{x}$, $\Tilde{\mathbf{x}} \in \mathbb{R}^{M\times 3J}$ are motions containing a sequence of $M$ poses (frames), each of which contains $J$ joint locations and $B$ bones. $BL$ computes the bone lengths in each frame. $q_{m,j}^k$ and $\Tilde{q}^k_{m,j}$ are the $k$th-order derivative of the $j$th joint in the $m$th frame in the clean sample and its adversary respectively. $k\in[0,2]$. This is because a dynamical system can be represented by a collection of derivatives at different orders. For human motions, we empirically consider the first three orders: position, velocity and acceleration. High-order information can also be considered but would incur extra computation. $||\cdot||_p$ is the $\ell_p$ norm. We set $p=2$ but other values are also possible. Overall, the first term is bone-length energy and the second one is motion dynamics energy. Both energy terms together define a distance function centered at a clean data $\mathbf{x}$. This distance function helps to quantify how likely an adversarial sample near $\mathbf{x}$ is, so \cref{eq:simMeasure} describes the adversarial distribution near the motion manifold. 

\section{Experiments}
\subsection{Experiments on Image Classification}\label{sec:image_exp}
\subsubsection{Experimental Settings}
We employ three popular image datasets, i.e. CIFAR-10~\cite{krizhevsky2009learning}, CIFAR-100~\cite{krizhevsky2009learning} and STL-10~\cite{coates2011analysis}. Since our work is closely related to the Bayesian defense strategy, we choose Bayesian defense methods Adv-BNN~\cite{AdvBNN} and IG-BNN~\cite{doan2022bayesian} as baselines. To ensure a fair comparison, we adopted their default settings. Specially, we use the VGG-16 network on CIFAR-10 as the target network. On STL-10, we use the smaller ModelA network used in Adv-BNN and IG-BNN. In addition, unlike existing Bayesian defense methods which are only feasible on small networks~\cite{AdvBNN,doan2022bayesian}, BBC can also be used on larger networks such as WideResNets~\cite{zagoruyko2016wide} since our proposed \textit{post-train} Bayesian strategy can avoid heavy memory footprint. To demonstrate this, we compare the results on WideResNets with the state-of-the-art AT methods. Most AT works on CIFAR-10 and CIFAR-100 use either a WRN28-10 or a WRN34-10 network, so we choose WRN28-10 on CIFAR-10 and WRN34-10 on CIFAR-100. Finally, since AAA~\cite{chenadversarial} is also a post-train defense method, we choose it as another baseline. The training details of BBC are in Appendix C. 

\noindent{\textbf{Attack Setting.}} The defenses are assessed by several gradient-based attacks under the $l_{\infty}$ and $l_{2}$ setting. White-box attacks includes vanilla PGD~\cite{madry_towards_2018}, EoT-PGD~\cite{athalye2018obfuscated} and AutoAttack (\cref{sec:auto})~\cite{auto}, and black-box attack includes Bandits~\cite{Bandits}. Unless specified otherwise, we set the $l_{\infty}$-norm bounded perturbation size to 8/255 and the attack iterations to 20. We report clean accuracy (accuracy on benign data) and robustness (accuracy on adversarial examples).  

\subsubsection{Robustness under White-box Attacks}
\noindent{\textbf{Comparison with Bayesian Defense.}} 
Following the evaluation protocol in~\cite{AdvBNN,doan2022bayesian}, we combine Expectation-over-Transformation~\cite{athalye2018obfuscated} with PGD~\cite{madry_towards_2018} to develop a strong white-box $l_{\infty}$-EoT PGD. We set the perturbation budget to $\epsilon \in [0:0.07:0.005]$ and report the results for BBC, PGD-AT~\cite{madry_towards_2018} and other Bayesian defense in~\cref{tab:Bay_def}. 

Overall, BBC demonstrates better robustness compared with other defenses, particularly as the attack strength increases. Notably, even at the extreme perturbation of 0.07, BBC still retains 48.4\% robustness on CIFAR-10, outperforming the previous methods by 31.9\%. More importantly, the robustness improvements are `for free', meaning that BBC does not compromise accuracy and require retraining the model under the black-box defense setting. 
\begin{table}[!htb]
\vspace{-0.3cm}
\begin{center}
\caption{Comparing robustness(\%) under different EoT-PGD attack budget.}\label{tab:Bay_def}%
\vspace{-0.3cm}
\begin{tabular}{@{}llllll@{}}
\toprule
Model\&Data &  Defenses  & 0 & 0.035 &0.055 &0.07\\
\midrule
\multirow{5}{*}{VGG-16 on CIFAR-10} &None  & 93.6  & 0 &0 &0 \\
&PGD-AT  & 80.3  & 31.1 &15.5 &10.3 \\
&Adv-BNN   & 79.7  & 37.7 &16.3 &8.1  \\
&IG-BNN   & 83.6  & 50.2  &26.8 &16.9 \\
&BBC(Ours)   & \textbf{93.3}  & \textbf{79.1}  &\textbf{63.1} &\textbf{48.4}\\
\cmidrule{0-0}
\multirow{5}{*}{ModelA on STL-10} &None  & 78.5  & 0 &0 &0 \\
&PGD-AT  & 63.2  & 27.4 &12.8 &7.0 \\
&Adv-BNN   & 59.9  & 31.4 &16.7 &9.1  \\
&IG-BNN   & 64.3  & \textbf{48.2}  &34.9 &27.3 \\
&BBC(Ours)   & \textbf{78.2}  & 43.3  &\textbf{41.2} &\textbf{39.3}\\
\bottomrule
\end{tabular}
\end{center}
\vspace{-0.3cm}
\end{table}

\noindent{\textbf{Comparisons with AT Methods.}} 
We select several AT methods, including the popular PGD-AT~\cite{madry_towards_2018}, TRADES~\cite{TRADES}, MART~\cite{MART}, FAT~\cite{zhang2020attacks}, LBGAT~\cite{LBGAT}, and the latest LAS-AT~\cite{LAS-AT} and AWP~\cite{AWP}. Moreover, we compare our method with a random-based defense RAT~\cite{rat}, which adds random noise into deterministic weights. All AT methods use the WRN34-10 network. We choose the PGD attack~\cite{madry_towards_2018} for evaluation. The results on CIFAR-10 and CIFAR-100 are reported in \cref{tab:comparsion_AT}, where BBC outperforms the AT methods by a large margin on both benign and robust performance. In addition, BBC simultaneously keeps similar high accuracy on adversarial samples and clean data, demonstrating the effectiveness of a full Bayesian treatment on the clean data, the adversarial samples and the classifier.
\begin{table}[!htb]
\vspace{-0.3cm}
\begin{center}
\caption{Comparing robustness(\%) on CIFAR-10 and CIFAR-100.}\label{tab:comparsion_AT}%
\vspace{-0.5cm}
\resizebox{1\linewidth}{!}{
\begin{tabular}{@{}lcccccc@{}}
\toprule
& \multicolumn{3}{@{}c@{}}{CIFAR-10} & \multicolumn{3}{@{}c@{}}{CIFAR-100}\\\cmidrule{2-4} \cmidrule{5-7}
Method  & Clean  &PGD-20 &PGD-50 & Clean  &PGD-20 &PGD-50\\
\midrule
ST        & 96.2    &0  &0 & 80.4   &0 &0\\
PGD-AT    & 85.2   &55.1 &54.9 &60.9 &31.7 &31.5\\
TRADES  & 85.7 & 56.1 &55.9 &58.6 &28.7 &26.6\\
MART &84.2 &58.6 &58.1 &60.8 &26.4 &25.8\\
LAS-AT  & 87.7 & 60.2 &59.8 &64.9 &36.4 &36.1\\
AWP  & 85.6 & 58.1 &57.9 &60.4 &33.9 &33.7\\
FAT  & 88.0 & 49.9 &48.8 &- &- &-\\
LBGAT  & 88.2 & 54.7 &54.3  &60.6 &34.8 &34.6\\ 
RAT & 86.1 & 61.4 &-  &64.7 &35.7 &-\\ 
BBC(Ours)    & \textbf{94.7}   & \textbf{93.9} & \textbf{93.7} &\textbf{74.4}
&\textbf{70.4} &\textbf{69.5}\\
\bottomrule
\end{tabular}}
\end{center}
\vspace{-0.3cm}
\end{table}

\noindent{\textbf{Comparisons with Robustness Model Using Extra Data.}}
It has been observed that the use of data augmentation and large models can improve robust generalization~\cite{gowal2020uncovering}. Hence we compare BBC with \cite{gowal2020uncovering}, which uses a large model, 500K additional unlabeled images extracted from 80M-Ti Million Tiny Images (80M TI) dataset~\cite{torralba200880} and other carefully designed experimental suites to considerably progress the state-of-the-art performance on multiple robustness benchmarks~\cite{croce2020robustbench}. We follow the evaluation protocol in \cite{gowal2020uncovering} to use PGD-40 under $l_{\infty}$ norm-bounded perturbations of size $\epsilon=8/255$ and $l_2$ norm-bounded perturbations of size $\epsilon=128/255$. As shown in~\cref{tab:augmentation}, BBC sets a new state-of-the-art on benign and PGD robust evaluation with no extra data, tiny increased model capacity, and no retraining. 
\begin{table}[!htb]
\vspace{-0.5cm}
\begin{center}
\caption{Comparing of BBC with \cite{gowal2020uncovering}.}\label{tab:augmentation}%
\vspace{-0.3cm}
\begin{tabular}{@{}lllll@{}}
\toprule
Data \& Models &  Norm  & Extra data & Clean &Robust\\
\midrule
CIFAR-10 & & & &\\
\cite{gowal2020uncovering}(WRN28-10)& $l_{\infty}$  & 80M TI  & 89.5   & 64.1 \\
\cite{gowal2020uncovering}(WRN70-16)& $l_{\infty}$  & 80M TI  & 91.1   & 67.2 \\
BBC(WRN28-10) &$l_{\infty}$ &None &94.7 &93.7 \\
\cite{gowal2020uncovering}(WRN70-16)& $l_2$  & 80M TI  & 94.7   & 82.2 \\
BBC(WRN28-10) &$l_2$ &None &94.7 &93.1 \\
\hline
CIFAR-100 & & & &\\
\cite{gowal2020uncovering}(WRN70-16)& $l_{\infty}$  & 80M TI  & 69.2   & 39.0 \\
BBC(WRN28-10) &$l_{\infty}$ &None & 74.4 &69.5 \\
\bottomrule
\end{tabular}
\end{center}
\vspace{-0.5cm}
\end{table}

\subsubsection{Robustness under Black-box Attack}
Under the black-box attack setting, we evaluate BBC along with AAA~\cite{chenadversarial} and AT~\cite{madry_towards_2018} defenses. Given AAA~\cite{chenadversarial} is specially designed to prevent score-based query attacks (SQA), we adapt it using SQA Bandits~\cite{Bandits}, which jointly leverages a time and data-dependent prior as a predictor of the gradient. We assess the defenses using both $l_{\infty}$ and $l_2$ threat models, and adopt the default attack settings as described in~\cite{Bandits}. Since \cite{chenadversarial} and \cite{Bandits} do not report results for AT and AAA on CIFAR-10 against $l_2$-Bandits, we only evaluate our method against $l_2$-Bandits in \cref{tab:bandits}. In comparison with AT, both AAA and BBC can improve robustness without hurting accuracy. But unlike AAA, whose robustness shows a big gap with accuracy (-16.4\%), the gap of BBC is much smaller (-2.6\%) across both $l_{\infty}$ and $l_2$ threat models. Again, this demonstrates the better robustness and accuracy of BBC.
\begin{table}[h]
\vspace{-0.3cm}
\begin{center}
\caption{Robustness(\%) under Bandits attack on CIFAR-10(@query=100/2500).}\label{tab:bandits}%
\vspace{-0.2cm}
\begin{tabular}{@{}lllll@{}}
\toprule
 &    &  &\multicolumn{2}{@{}c@{}}{Bandits}\\\cmidrule{4-5}
Method &Norm &Clean &@100 &@2500 \\
\midrule
ST &$l_{\infty}$=8/255 &96.2 &69.9 &41.0\\
AT &$l_{\infty}$=8/255 &87.0 &83.6 &76.3\\
AAA &$l_{\infty}$=8/255 &\textbf{94.8} &80.9 &78.4\\
BBC(Ours) &$l_{\infty}$=8/255 &94.7 &\textbf{93.7} &\textbf{93.4}\\\cmidrule{1-2}
ST &$l_{2}$ &96.2 &1.3 &0\\
BBC(Ours) &$l_{2}$ &94.7 &\textbf{93.8} &\textbf{92.1}\\
\bottomrule
\end{tabular}
\end{center}
\vspace{-0.7cm}
\end{table}

\subsection{Experiments on S-HAR}
\subsubsection{Experimental Settings}
\label{sec:setting}
We briefly introduce the experimental settings here, and the details are in Appendix C.

\noindent{\textbf{Datasets and Classifiers.}} We choose three widely adopted benchmark datasets in HAR: HDM05~\cite{muller_documentation_2007}, NTU60~\cite{shahroudy_ntu_2016} and NTU120~\cite{liu_ntu_2020}. For base classifiers, we employ four recent classifiers: ST-GCN~\cite{yan_spatial_2018}, CTR-GCN~\cite{chen_channel-wise_2021}, SGN~\cite{zhang_semantics-guided_2020} and MS-G3D~\cite{liu_disentangling_2020}. Since these classifiers do not have the same setting (e.g. data needing sub-sampling~\cite{zhang_semantics-guided_2020}), we unify the data format. For NTU60 and NTU120, we sub-sample the frames to 60. For HDM05, we divide the data into 60-frame samples~\cite{wang_understanding_2021}. Finally, we retrain the classifiers following their original settings. 

\noindent{\textbf{Attack Setting.}} We employ recent attackers designed for S-HAR: SMART ($l_2$ attack)~\cite{wang_understanding_2021}, CIASA ($l_{\infty}$ attack)~\cite{liu_adversarial_2020} and BASAR ($l_2$ attack)~\cite{diao_basarblack-box_2021}. Further, we use the untargeted attack, which is the most aggressive setting. In \cite{wang_understanding_2021}, SMART uses a learning rate of 0.005 and a maximum of 300 iterations. Note that we conduct a higher number of attack iterations than SMART reported in their papers, as increased iterations result in more aggressive attacks. We use learning rate of 0.01 and increase the maximum iterations of SMART to 1000 to evaluate the effectiveness of our method against strong attacks. CIASA~\cite{liu_adversarial_2020} uses a learning rate of 0.01 and does not report the exact maximum iterations. Similarly, we maintain the default learning rate and increase iterations to 1000 on HDM05. Since running 1000-iteration CIASA on NTU 60/120 is prohibitively slow (approximately 1 month on one Nvidia RTX 3090 GPU), we employ 100-iteration CIASA. For other settings, we follow their default settings~\cite{wang_understanding_2021,liu_adversarial_2020}. We use the same iterations for BASAR as in their paper~\cite{diao_basarblack-box_2021}. Same as the evaluation metric used in image experiments, we report clean accuracy and robustness. 

\noindent{\textbf{Defense Setting.}} To our best knowledge, BBC is the first black-box defense for S-HAR. So there is no method for direct comparison. There is a technical report~\cite{zheng_towards_2020} which is a simple direct application of randomized smoothing (RS)~\cite{Cohen19_ICML}. We use it as one baseline. AT~\cite{madry_towards_2018} has recently been attempted on HAR~\cite{BASAR_journal,tanaka2021adversarial}, so we use it as a baseline SMART-AT which employs SMART as the attacker. We also employ another two baseline methods TRADES~\cite{TRADES} and MART~\cite{MART}, which are the state-of-the-art defense methods on images. Besides, we compare BBC with the state-of-the-art randomized defense RAT~\cite{rat}. We employ perturbations budget $\epsilon = 0.005$ for training AT methods~\cite{madry_towards_2018,MART,TRADES} and compare other $\epsilon$ settings in \cref{sec:evaluation}.

\noindent{\textbf{Computational Complexity.}}  We use 20-iteration attack for training AT-based methods, since more iterations incur much higher computational overhead than BBC, leading to unfair comparison. We compare the training time of BBC with other defenses on all datasets and the results are reported in Appendix A. Since BBC does not need to re-train the target model, it reduces training time (by 12.5\%-70\%) compared with the baselines.

\subsubsection{Robustness under White-box Attacks}
\label{sec:evaluation}

\begin{table*}[ht]
\vspace{-0.3cm}
\begin{center}
\begin{minipage}{\textwidth}
\caption{Clean accuracy(\%) and robustness(\%) on HDM05 (top), NTU60 (middle) and NTU120 (bottom).}
\vspace{-0.3cm}
\label{tab:comparisons_hdm05_ntu60_ntu120}
\resizebox{1\linewidth}{!}{
\begin{tabular}{l ccc|ccc|ccc|ccc}
\toprule
\multirow{2}{*}{\small{HDM05}} & \multicolumn{3}{c }{ST-GCN} & \multicolumn{3}{c }{CTR-GCN} & \multicolumn{3}{c }{SGN} & \multicolumn{3}{c }{MS-G3D}\\\cmidrule{2-4}\cmidrule{5-7}\cmidrule{8-10}\cmidrule{11-13}
& \small{Clean} & \small{SMART} & \small{CIASA} & \small{Clean} & \small{SMART} & \small{CIASA} & \small{Clean} & \small{SMART} &\small{CIASA} &\small{Clean} & \small{SMART}& \small{CIASA}\\
\midrule
\small{Standard}& 93.2 & 0 & 0 & 94.2 & 9.9 & 8.8 & 94.2 & 1.9 & 0.4 & 93.8 & 3.0 & 3.9\\

\small{SMART-AT}& 91.9 & 10.5 & 8.6 & 93.0 & 22.6 & 18.3 & 93.3 & 3.1 & 2.5 &92.8 & 28.4 & 21.3\\

\small{RS} & 92.7 & 3.6 & 2.9 & 92.1 & 17.1 & 18.3 & 92.8 & 7.9 & 1.5 & 93.0 & 4.7 & 5.0\\

\small{MART} & 91.1 & 17.5 & 15.0 & 91.5 & 29.6 & 20.4 & 93.8 & 2.7 & 1.5 & 91.5 & 39.9 & 18.2\\

\small{TRADES} & 91.5 & 18.8 & 13.7 & 92.8 & 24.3 & 23.2 & 92.3 & 3.4 & 0.1 & 90.0 & 39.8 & 41.3\\

\small{RAT} & 88.1 & 16.0 & 12.4 & 90.4 & 30.9 & 24.2 & 92.3 & 15.1 & 8.3 & 93.2 & 61.9 & 61.9\\

\small{BBC(Ours)} & \textbf{93.0} & \textbf{35.8} & \textbf{30.3} & \textbf{93.2} & \textbf{32.7} & \textbf{31.7} & \textbf{94.7} & \textbf{69.3} & \textbf{68.6} & \textbf{93.6} & \textbf{74.7} & \textbf{74.4}\\         
\bottomrule
\end{tabular}}

\resizebox{1\linewidth}{!}{
\begin{tabular}{l ccc|ccc|ccc|ccc}
\toprule
\multirow{2}{*}{\small{NTU60}} & \multicolumn{3}{c }{ST-GCN} & \multicolumn{3}{c }{CTR-GCN} & \multicolumn{3}{c }{SGN} & \multicolumn{3}{c }{MS-G3D}\\\cmidrule{2-4}\cmidrule{5-7}\cmidrule{8-10}\cmidrule{11-13}
& \small{Clean} & \small{SMART} & \small{CIASA} & \small{Clean} & \small{SMART} & \small{CIASA} & \small{Clean} & \small{SMART} &\small{CIASA} &\small{Clean} & \small{SMART}& \small{CIASA}\\
\midrule         
 \small{Standard}& 76.8 & 0\ & 0.5 & 82.9 & 0 & 0 & 86.2 & 0 & 2.3 & 89.4 & 0.3 & 2.0 \\
 
 \small{SMART-AT} & 72.8 & 0 & 10.6 & \textbf{83.7} & 0 & 19.6 & 83.3 & 0 & 10.5 & 87.8 & 0 & 40.4\\
 
 \small{RS} & 75.9 & 0 & 4.0 & 82.7 & 0 & 6.7 & 83.0 & 0 & 5.9 & 88.1 & 0 & 10.0 \\
 
 \small{MART} & 71.9 & 0 & 14.6 & 80.3 & 0 & 16.3 & 83.2 & 0 & 14.0 & 85.4 & 0 & 42.2\\
 
 \small{TRADES} & 71.4 & 0 & 18.7 & 79.6 & 0 & 18.5 & 82.3 & 0 & 19.3 & 85.2 & 0 & 43.8\\

 \small{RAT} & 74.5 & 4.3 & \textbf{25.4} & 80.0 & 5.6 & \textbf{32.6} & 83.0 & 1.7 & 26.4 & 84.8 & 1.8 & 47.1\\
 
 \small{BBC(Ours)} & \textbf{76.5} & \textbf{28.3} & 22.0 & 82.8 & \textbf{22.2} & 30.8 & \textbf{86.1} & \textbf{51.6} & \textbf{56.5} & \textbf{88.8} & \textbf{60.1} & \textbf{58.4}\\
\bottomrule
\end{tabular}}
\resizebox{1\linewidth}{!}{
\begin{tabular}{l ccc|ccc|ccc|ccc}
\toprule
\multirow{2}{*}{\small{NTU120}} & \multicolumn{3}{c }{ST-GCN} & \multicolumn{3}{c }{CTR-GCN} & \multicolumn{3}{c }{SGN} & \multicolumn{3}{c }{MS-G3D}\\\cmidrule{2-4}\cmidrule{5-7}\cmidrule{8-10}\cmidrule{11-13}
& \small{Clean} & \small{SMART} & \small{CIASA} & \small{Clean} & \small{SMART} & \small{CIASA} & \small{Clean} & \small{SMART} &\small{CIASA} &\small{Clean} & \small{SMART}& \small{CIASA}\\
\midrule                  
 \small{Standard}& 68.3 & 0 & 0.6 & 74.6 & 0 & 0.3 &74.2 &0 &0.6 &84.7 &0.4 &1.9\\
 
 \small{SMART-AT} & 67.3 & 0 & 10.2 & \textbf{75.9} & 0 & 8.7 & 71.3 & 0 & 3.8 & 81.9 & 0.5 & 29.8 \\
 
 \small{RS} & 66.8 & 0 & 3.0 & 74.0 & 0 & 3.6 & 71.4 & 0 & 1.0& 82.2 & 0.1 & 1.3\\
 
 \small{MART} & 58.4 & 0.1 & 11.3 & 70.5 & 0.1 & 13.8 & 70.1 & 0.1 & 9.8 & 78.9 & 0.1 & 33.4\\
 
 \small{TRADES} & 61.4 & 0.2 & 10.6 & 72.0 & 0 & 13.4 & 69.4 & 0 & 11.3 & 79.0 & 0 & 35.3\\

 \small{RAT} & 62.8 & 0.2 & 13.7 & 72.0 & 0.6 & \textbf{20.4} & 67.1 & 0 & 14.6 & 77.3 & 1.8 & 39.2\\
 
 \small{BBC(Ours)} & \textbf{68.3} & \textbf{10.6} & \textbf{15.6} & 74.6 & \textbf{10.7} & 16.5 & \textbf{73.5} & \textbf{32.0} & \textbf{46.2} & \textbf{84.7} & \textbf{50.5} & \textbf{50.9}\\
\bottomrule
\end{tabular}}
\end{minipage}
\end{center}
\vspace{-0.5cm}
\end{table*}

We show the results in \cref{tab:comparisons_hdm05_ntu60_ntu120}. First, BBC does not severely compromise the clean accuracy across models and data. Its accuracy is within a small range (+0.6/-0.9\%) from that of the standard training, in contrast to the often noticeable accuracy drop in other methods. Next, BBC outperforms the baseline methods under almost all scenarios (classifiers vs. datasets vs. attackers). The only exception is RAT~\cite{rat}, which outperforms BBC in 3 out of 24 attack scenarios, at the cost of compromising the accuracy. However, BBC still overall outperforms RAT, especially under more aggressive attacks, such as defending against extreme SMART-1000. Besides, RAT needs to re-train the target model, hence consuming more training time than BBC, as reported in \cref{sec:setting}. Overall, BBC can significantly improve the adversarial robustness and eliminate the accuracy-robustness trade-off.

\begin{figure}[!htb]
    \centering
    \includegraphics[width=0.95\linewidth]{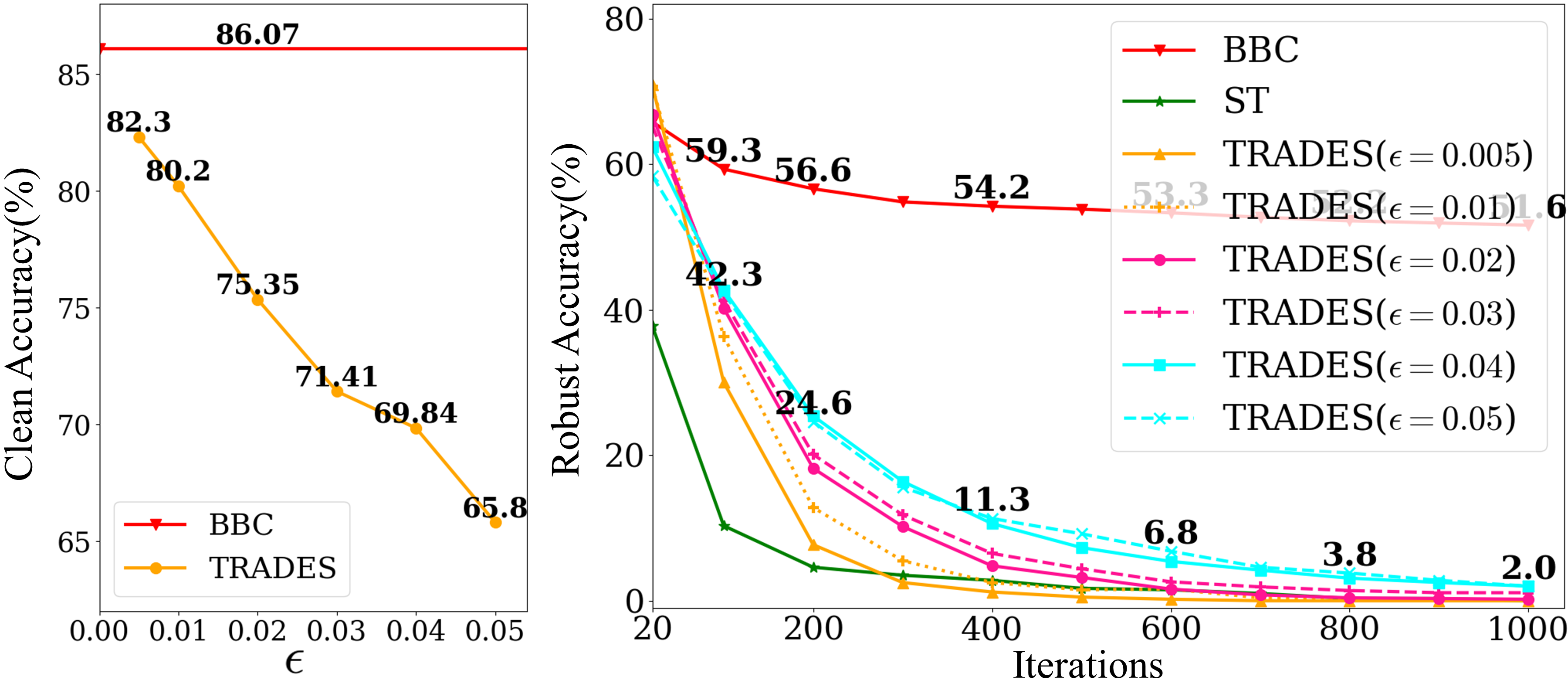}
    \caption{Comparisons with TRADES with different perturbation budget ($\epsilon$) on NTU60 with SGN. (a): standard accuracy vs. $\epsilon$; (b): robustness against SMART with 20 to 1000 iterations.}
    \label{fig:acc_epsilon}
\end{figure}

\noindent{\textbf{Comparison with other AT methods.}} In~\cref{tab:comparisons_hdm05_ntu60_ntu120}, all baseline methods are worse than BBC, sometimes completely fail, e.g. failing to defend against SMART-1000 in large datasets (NTU 60 and NTU 120). After investigating their defenses against SMART from iteration 20 to 1000 in \cref{fig:attack_strength} (full results reported in Appendix A), we found the key reason is the baseline methods overly rely on the aggressiveness of the adversaries sampled during training. To verify it, we increase the perturbation budget $\epsilon$ from 0.005 to 0.05 during training in TRADES, and plot their clean accuracy \& robustness vs. $\epsilon$ in~\cref{fig:acc_epsilon}. Note that BBC does not rely on a specific attacker. We find TRADES is highly sensitive to $\epsilon$ values: larger perturbations in adversarial training improve the defense (albeit still less effective than BBC), but harm the standard accuracy (\cref{fig:acc_epsilon}(a)). Further, sampling adversaries with more iterations (e.g. 1000 iterations) during AT may also improve the robustness (still worse than BBC~\cref{fig:acc_epsilon}(b)) but is prohibitively slow, while BBC requires much smaller computational overhead (see \cref{sec:setting}). 

\begin{figure*}[!htb]
    \centering
    \includegraphics[width=0.95\linewidth]{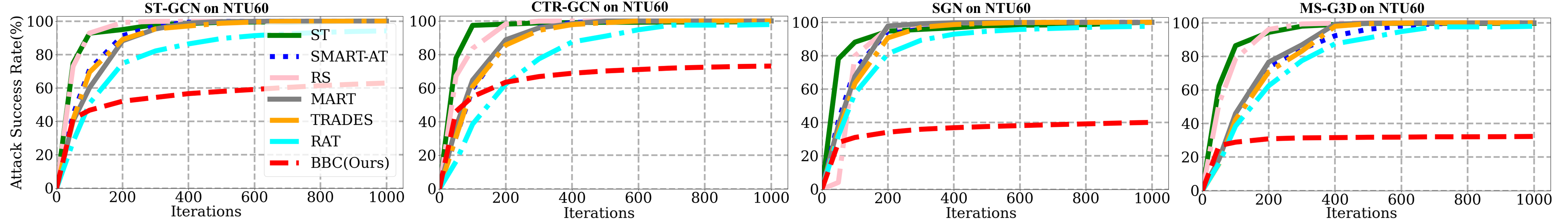}
    \caption{The attack success rate vs. attack strength curves against SMART on NTU 60. For each subplot, the abscissa axis is iterations while the ordinate axis is the attack success rate(\%). ST means standard training.}
    \vspace{-0.3cm}
    \label{fig:attack_strength}
    \vspace{-0.2cm}
\end{figure*}
\subsubsection{Robustness under Black-box Attacks}
\label{sec:evaluation_black}
Black-box attack in S-HAR is either transfer-based~\cite{wang_understanding_2021} or decision-based~\cite{diao_basarblack-box_2021}. However, existing transfer-based attacks (SMART and CIASA) are highly sensitive to the chosen surrogate and the target classifier. According to our preliminary experiments (see Appendix A), transfer-based SMART often fails when certain models are chosen as the surrogate, which suggests that transfer-based attack is not a reliable way of evaluating defense in S-HAR. Therefore we do not employ it for evaluation. BASAR is a decision-based approach, which is truly black-box and has proven to be far more aggressive~\cite{diao_basarblack-box_2021}. We employ its evaluation metrics, i.e. the averaged $l_2$ joint position deviation ($l$), averaged $l_2$ joint acceleration deviation ($\Delta a$) and averaged bone length violation percentage ($\Delta B/B$), which all highly correlate to the attack imperceptibility. We randomly select samples following~\cite{diao_basarblack-box_2021} for attack. The results are shown in~\cref{tab:basar}. BBC can often reduce the quality of adversarial samples, which is reflected in $l_2$, $\Delta a$ and $\Delta B/B$. The increase in these metrics means severer jittering/larger deviations from the original motions, which is very visible and raises suspicion. We show the visual examples in Appendix A.
\begin{table}[htb]
\vspace{-0.3cm}
\begin{center}
\begin{minipage}{\linewidth}
\caption{Untargeted attack on HDM05 (top), NTU60 (middle) and NTU120 (bottom) from BASAR. xxx/xxx is pre/post BBC results.}
\vspace{-0.2cm}
\resizebox{1\linewidth}{!}{
\begin{tabular}{l l l l l}
\toprule
& \small{STGCN} & \small{CTRGCN} & \small{SGN} & \small{MSG3D} \\
\midrule
HDM05 & & & & \\
$l$ $\uparrow$ & 0.77/\textbf{0.82} & 0.67/\textbf{0.79} & 0.84/\textbf{1.05} & 0.20/\textbf{0.28} \\
$\Delta a$ $\uparrow$ & 0.21/\textbf{0.22} & 0.14/0.14 & 0.05/\textbf{0.07} & 0.086/\textbf{0.095} \\
$\Delta B/B$ $\uparrow$ & 0.42\%/\textbf{0.77\%} & 0.80\%/\textbf{0.94\%} & 1.1\%/\textbf{1.5\%} & 1.1\%/\textbf{1.2\%} \\
NTU60 & & & & \\
$l$ $\uparrow$ & 0.03/\textbf{0.05} & 0.05/\textbf{0.06} & 0.06/\textbf{0.08} & 0.09/0.09 \\
$\Delta a$ $\uparrow$ & 0.015/\textbf{0.017} & 0.02/\textbf{0.03} & 0.003/\textbf{0.004} & 0.03/\textbf{0.04} \\
$\Delta B/B$ $\uparrow$ & 4.2\%/\textbf{4.8\%} & 6.5\%/\textbf{7.4\%} & 1.3\%/\textbf{1.7\%} & 8.9\%/\textbf{11.0\%} \\
NTU120 & & & & \\
$l$ $\uparrow$ & 0.03/\textbf{0.04} & 0.04/\textbf{0.06} & 0.087/\textbf{0.103} & 0.06/\textbf{0.08} \\
$\Delta a$ $\uparrow$ & 0.015/\textbf{0.018} & 0.019/\textbf{0.022} & 0.005/\textbf{0.006} & 0.02/\textbf{0.03} \\
$\Delta B/B$ $\uparrow$ & 4.0\%/\textbf{4.7\%} & 5.4\%/\textbf{5.6\%} & 2.3\%/\textbf{2.7\%} & 6.8\%/\textbf{9.0\%} \\
\bottomrule
\end{tabular}}
\label{tab:basar}
\end{minipage}
\end{center}
\vspace{-0.2cm}
\end{table}

\begin{table}[!htb]
\vspace{-0.3cm}
\begin{center}
\caption{Robustness against AutoAttack on CIFAR-10. Left: randomized transformation to inputs. Right: Adversarial detection via BBC.}\label{tab:auto_detect}%
\vspace{-0.3cm}
\resizebox{1\linewidth}{!}{
\setlength{\tabcolsep}{0.5 mm}{
\begin{tabular}{@{}l|ccc||cccccc@{}}
\toprule
& \multicolumn{3}{c||}{Randomized Transformation} &\multicolumn{3}{@{}c@{}}{Detection($l_{\infty}$=4/255)} & \multicolumn{3}{@{}c@{}}{Detection($l_{\infty}$=8/255)}\\
$\sigma$ &Clean &$l_{\infty}$=4/255 &$l_{\infty}$=8/255 & TPR$\uparrow$  &FPR$\downarrow$ &AUROC$\uparrow$ & TPR$\uparrow$  &FPR$\downarrow$ &AUROC$\uparrow$\\
\midrule
0.02 &91.5 & 26.5 &6.3    & 73.3\%    &11.0\%  &0.81 &63.2\%    &11.0\% &0.76\\
0.03 &85.2 &39.8 &13.3   &82.4\%    &19.4\% &0.81 &66.4\% &19.4\% &0.74\\
\bottomrule
\end{tabular}}}
\end{center}
\vspace{-0.5cm}
\end{table}


\subsection{Evaluation under AutoAttack}
\label{sec:auto}
We evaluate BBC against AutoAttack~\cite{auto} separately, because AutoAttack is a strong ensemble attack, consisting of three types of white-box attacks APGD-CE, APGD-DLR, FAB~\cite{fab} and one black-box attack Square~\cite{square}, which has been shown to be more aggressive than many attack methods and therefore suitable for testing the robustness of the defender. We initially find that it is difficult for BBC to defend against AutoAttack under the same settings as before. Therefore, we propose two variants of BBC that can defend AutoAttack. 

\subsubsection{Randomization of Adversarial Examples}
AutoAttack involves exhaustive sampling, while BBC relies on sampling of potential adversarial samples near the data manifold. Since the manifold dimension is high, there can be void spaces where samples are not drawn by BBC and in these void spaces AutoAttack could still find adversarial examples. Looking closely at \cref{eq:SGLD_xtilde}, the distance between $\Tilde{\mathbf{x}}$ and $\mathbf{x}$ is governed by $\epsilon$. In other words, samples can be more concentrated in areas that are a certain distance (controlled by $\epsilon$) away from the manifold, so that the areas closer to the manifold become void, where AutoAttack can still find adversarial samples. A naive solution to this problem would be to employ a range of $\epsilon$ values during sampling, but this would make the training slower.

To tackle this problem, we do not change the training process but add Gaussian noise to adversarial examples before feeding them into the model. The Gaussian noise is drawn from $\mathcal{N}(0, \sigma^2 \mathbf{I})$ where $\sigma$ is the standard deviation. As shown in \cref{tab:auto_detect} (left), the addition of random noise to adversarial examples enables BBC to move them to the learned adversarial distribution, leading to good defense results. This proves our previous speculation about possible void spaces near the manifold, and also leads us to a new defense setting below.

Further motivated by this observation, we design a simple but effective adversarial detection method to further enhance BBC against AutoAttack: given a test sample $\Tilde{\mathbf{x}}$, we detect it as adversarial if $\argmax F(\Tilde{\mathbf{x}}+\eta)) \neq \argmax F(\Tilde{\mathbf{x}}))$, where $F(\cdot)$ is the output of BBC and $\eta$ is the Gaussian noise. We measure TPR (true positive rate) which indicates the percentage of adversarial examples that are successfully detected, FPR (false positive rate) which indicates the percentage of clean examples that are misclassified, and the area under the receiver operating characteristic curve (AUROC). The results are presented in \cref{tab:auto_detect} (right). The high AUROC value indicates a large proportion of adversarial examples are successfully detected while maintaining a low misclassification rate. This result demonstrates the robustness of BBC against AutoAttack.

\subsubsection{Energy-based Pretraining}
As a post-train black-box defense, a fundamental advantage of BBC is that it does not modify latent features through re-training. However, we speculate that the pre-trained latent features from clean data can be intrinsically vulnerable to strong adversarial attacks, which cannot be completely eliminated by BBC. This is because, while BBC ideally considers the whole adversarial distribution, it is practically impossible to account for every adversarial case. This vulnerability is intrinsic due to the distribution of the learned features in pre-training. In contrast, AutoAttack, by taking a wide range of attacks into consideration, can always find the most vulnerable adversarial subspace that BBC may overlook.

To verify this speculation, we allow BBC to interfere with the pre-training process, by replacing the standard pre-trained model trained on clean data with a robustly pre-trained model, so that the post-train strategy can further improve the robustness performance. But rather than choosing other robust training methods such as adversarial training or randomized smoothing, we use a simplified version of BBC for pre-training. Specifically, we employ JEM~\cite{grathwohl_your_2020} which is an energy-based method that maximizes the joint probability $p_{\theta}(\mathbf{x}, y) = \frac{exp(g_{\theta}(\mathbf{x})[y])}{Z(\theta)}$. It is a simplified version of BBC due to that it does not consider the adversarial distributions or the model distribution. We pretrain the architecture based on WRN28-10 using JEM, then apply BBC. The results are reported in~\cref{tab:jem}. In summary, BBC can fortify pre-trained models lacking robustness, such as those resulting from standard training. Furthermore, for robust pre-trained models, BBC contributes to an additional enhancement in their robustness.

\begin{table}[htb]
\vspace{-0.3cm}
\begin{center}
\caption{Robustness(\%) on CIFAR-10 using JEM as the pre-trained model.}\label{tab:jem}
\vspace{-0.3cm}
\resizebox{1\linewidth}{!}{
\begin{tabular}{ccccccc}
\toprule
Model   & Clean & APGD-CE & APGD-DLR & FAB  & SQUARE & AutoAttack \\
\midrule
JEM     & 92.9  & 6.6     & 12.5     & 10.9 & 18.7   & 5.5        \\
JEM+BBC & 92.1  & 85.4    & 22.7     & 74.4 & 44.8   & 15.5      \\
\bottomrule
\end{tabular}}
\end{center}
\vspace{-0.8cm}
\end{table}

\subsection{Gradient Not Obfuscated}
As pointed in \cite{athalye2018obfuscated}, some defense methods rely on obfuscated gradients which can be circumvented easily. However, BBC is not the case. The key reason for BBC’s robustness is its ability to model the clean-adversarial joint data distribution and turn a definitive classification boundary into a boundary distribution, in contrast to obfuscated-gradient methods which hide the gradient for one deterministic boundary. Specifically, according to the criteria in \cite{athalye2018obfuscated}, BBC's gradient is by definition not shattered, stochastic or exploding/vanishing.

Further, empirical evidence shows that BBC does not rely on obfuscated gradients. First, iterative attack has a higher attack success rate than one-step attack in BBC. Iterative attacks are strictly stronger than one-step attack. If one-step attacks achieves higher attack success rate than iterative methods, it indicates the defense might rely on obfuscated gradients. The robustness of BBC on FGSM attack (90.5\%) is better than EOT-PGD-20 (79.1\%) on CIFAR10 using VGG-16. This indicates that one-step FGSM has lower attack success rate than the iterative EOT-PGD-20, demonstrating that BBC does not rely on obfuscated gradients. Second, larger distortion attack causes lower robustness in BBC (\cref{tab:Bay_def}). According to \cite{athalye2018obfuscated}, if larger distortion attacks do not increase the attack success rate, \ie lower robustness, it is an indication of possible gradient obfuscation. This is not the case in BBC. Last, our proposed defense is robust to AutoAttack and adaptive attacks such as FAB (\cref{tab:jem}), which has demonstrated its robustness to the phenomenon of gradient obfuscation~\cite{fab}.

More importantly, we follow \cite{athalye2018obfuscated} to verify gradient obfuscation by using Expectation-over-Transformation (EoT)~\cite{athalye2018obfuscated}. For image classification, we adopt EoT-PGD as mentioned earlier, and the results are reported in \cref{tab:Bay_def}. For S-HAR, we deploy an adaptive attack called EoT-SMART: in each step, we estimate the expected gradient by averaging the gradients of multiple randomly interpolated samples. \cref{tab:EOT} shows that EoT-SMART performs only slightly better than SMART, demonstrating that BBC does not rely on obfuscated gradients.

\begin{table}[htb]
\vspace{-0.3cm}
\begin{center}
\caption{Robustness(\%) against EoT-SMART. $\pm$xx) means the robustness difference with SMART.}
\vspace{-0.3cm}
\begin{tabular}{@{}lllll@{}}
\toprule
BBC & ST-GCN & CTR-GCN  &SGN  &MS-G3D  \\ 
\midrule
HDM05 & 35.1 (-0.7) &29.5 (-3.2)  &68.9 (-0.4)  & 71.8 (-2.9) \\ 
NTU 60 & 28.2 (-0.1) &22.3 (+0.1)  &50.0 (-1.6)  & 58.4 (-0.0) \\ 
\bottomrule
\end{tabular}
\label{tab:EOT}
\end{center}
\vspace{-0.5cm}
\end{table}




\subsection{Ablation Study}
We report the main ablation studies here, more results are presented in Appendix B.
\label{sec:ablation}
\subsubsection{Number of Appended Models.} Although BNNs theoretically require sampling of many models for inference, in practice, we find a small number of models suffice. To show this, we conduct an ablation study on the number of appended models (N in \cref{eq:ptBayesianClassifier}). For image classification, we report the robustness against APGD~\cite{auto}, which is a more competitive adversary than vanilla PGD attack. As shown in \cref{tab:num}, with $N$ increasing, BBC significantly lowers the attack success rates, which shows the Bayesian treatment of the model parameters is able to greatly increase the robustness. Further, when N $>$ 5, there is a diminishing gain in robustness but with increased computation. This performance is consistent across different models and data types, so we use N=5 by default. 

\begin{table}[h]
\vspace{-0.3cm}
\begin{center}
\caption{Robustness(\%) under different number of appended models.}\label{tab:num}%
\vspace{-0.3cm}
\begin{tabular}{@{}lllll@{}}
\toprule
&\multicolumn{2}{@{}c@{}}{CIFAR-10(WRN28-10)}  &\multicolumn{2}{@{}c@{}}{NTU60(SGN)}\\\cmidrule{2-3} \cmidrule{4-5}
Num&Clean &APGD &Clean &SMART \\
\midrule
1 &92.3 &15.9 &84.9 &3.1\\
3 &93.3 &68.9 &85.7 &36.9\\
5 &94.7 &88.7 &86.1 &51.6\\
7 &94.5 &89.2 &86.0 &62.4\\
\bottomrule
\end{tabular}
\end{center}
\vspace{-0.5cm}
\end{table}

\begin{figure}[!t]
    \centering
    \includegraphics[width=0.9\linewidth]{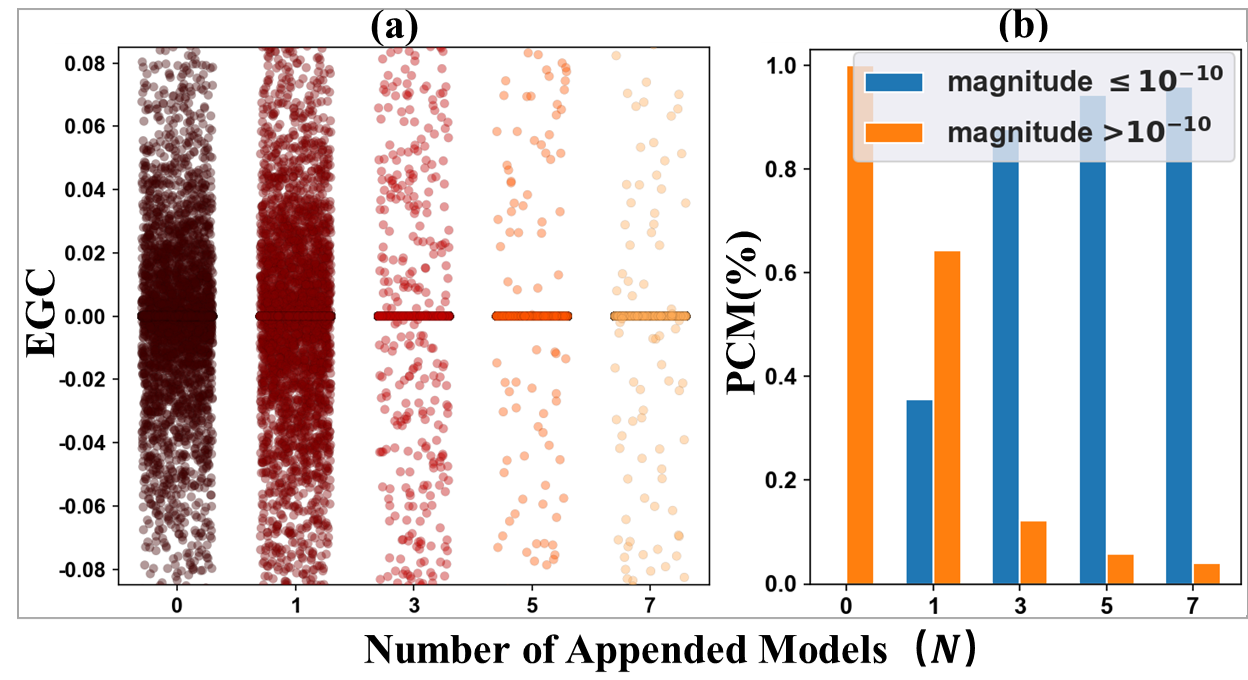}
    \vspace{-0.5cm}
    \caption{The components of the expected loss gradients of BBC on CIFAR-10 with WRN28-10. $N=0$ is standard training.  (a): the values of the expected gradient components(EGC); (b): the percentage of the component magnitude (PCM) above and below 10$^{-10}$.}
    \label{fig:loss_gradient_image}
    \vspace{-0.5cm}
\end{figure}

We further show why our \textit{post-train} Bayesian strategy is able to greatly increase the robustness. The classification loss gradient with respect to data is key to many attack methods. In a deterministic model, this gradient is computed on one model; in BBC, this gradient is averaged over all models, i.e. the expected loss gradient. Theoretically, with an infinitely wide network in the large data limit, the expected loss gradient achieves 0, which is the source of the good robustness of BNNs~\cite{Bortolussi_onTheRobustness_2022}. To investigate whether BBC's robustness benefits from the vanishing expected gradient, we randomly sample 500 images from CIFAR-10 and randomly count for 75000 loss gradient components in \cref{fig:loss_gradient_image}, which essentially show the empirical distribution of the component-wise expected loss gradient. As $N$ increases, the gradient components steadily approach zero, indicating a vanishing expected loss gradient which provides robustness~\cite{Bortolussi_onTheRobustness_2022}. BBC on motion data has similar results, which are reported in Appendix A.


\subsubsection{Joint Distribution of Data and Adversaries.} Other than the Bayesian treatment of models, BBC also benefits from the Bayesian treatment on the adversaries. To see its contribution, we plug-and-play our post-train Bayesian strategy to other AT methods which do not model the adversarial distribution. Specially, we keep the pre-trained classifiers intact and then take a post-train Bayesian treatment on TRADES(PB+TRADES). We compare it with BBC. As shown in~\cref{tab:ablation}, PB+TRADES only improves the robustness slightly on image data. On NTU60 with SGN as the base action classifier, PB+TRADES achieves 84.7\% clean accuracy and 34.7\% robustness against SMART. BBC still outperforms PB+TRADES by large margins, further improving accuracy by 1.3\% and robustness by 16.9\%. Note the major difference between BBC and PB+TRADES is whether to consider the full adversarial distribution, which shows the benefit of bringing the full adversarial distribution into the joint probability.
\begin{table}[h]
\vspace{-0.3cm}
\begin{center}
\caption{Ablation Study on CIFAR-10 with VGG-16 as the base classifier.}\label{tab:ablation}%
\vspace{-0.3cm}
\begin{tabular}{@{}lcccc@{}}
\toprule
Methods &  Clean  & PGD & APGD &EoT-PGD\\
\midrule
PB+TRADES &93.6 &10.7 &8.5 &3.9\\
BBC &93.3 &92.5 &89.1 &81.8\\
\bottomrule
\end{tabular}
\end{center}
\vspace{-0.8cm}
\end{table}

\section{Limitation and Discussion}
One limitation is that we need specific domain knowledge in instantiating $d(\mathbf{x}, \Tilde{\mathbf{x}})$ in \cref{eq:fullJointEnergy}. However, this is lightweight as manifold learning/representation is a rather active field and many methods could be used. BBC can potentially incorporate any manifold representation. Also, we assume that all adversarial samples are distributed closely to the data manifold, which is true for images~\cite{stutz2019disentangling} and skeletal motion~\cite{diao_basarblack-box_2021}, but not necessarily for other data.
Further, as a post-train black-box defense method, BBC aims not to change the latent features of the pre-trained models. This might require additional operations such as adding noise to the data to achieve good defense against strong attack methods e.g. AutoAttack. 

\section{Conclusions and Future Work}
To our best knowledge, we proposed a new post-train black-box defense framework and it is the first black-box defense for S-HAR. Our method BBC is underpinned by a new Bayesian Energy-based Adversarial Training framework, and is evaluated across various classifiers, datasets and attackers. Our method employs a post-train strategy for fast training and a full Bayesian treatment on clean data, their adversarial samples and the classifier, without adding much extra computational cost. In future, we will extend BBC to more data types, such as videos and graphs, by employing task/data specific $d(\mathbf{x}, \Tilde{\mathbf{x}})$ in \cref{eq:adDist}.




\bibliographystyle{./bibtex/IEEEtran}
\bibliography{./bibtex/IEEEabrv,references}

 




\begin{IEEEbiography}[{\includegraphics[width=1in,height=1.25in,clip,keepaspectratio]{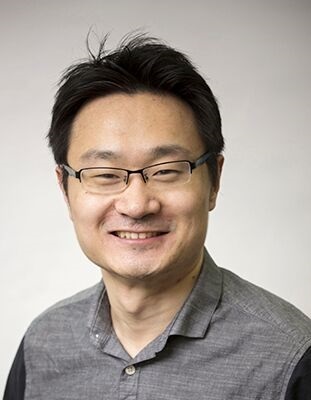}}]{He Wang}
is an Associate Professor at the Department of Computer Science, University College London and a Visiting Professor at the University of Leeds. He serves as an Associate Editor of Computer Graphics Forum. His research interest is mainly in computer graphics, computer vision and machine learning. He received his PhD from the University of Edinburgh.
\end{IEEEbiography}

\begin{IEEEbiography}[{\includegraphics[width=1in,height=1.25in,clip,keepaspectratio]{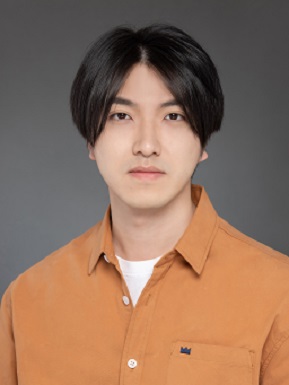}}]{Yunfeng Diao} is a Lecturer in the School of Computer Science and Information Engineering, Hefei University of Technology, China. He received his PhD from Southwest Jiaotong University, China. His current research interests include computer vision and the security of machine learning, particularly in adversarial examples and adversarial robustness.
\end{IEEEbiography}


\end{document}